\documentclass[runningheads]{llncs}
\usepackage[T1]{fontenc}
\usepackage{graphicx}
\usepackage{amssymb}
\usepackage{amsmath}
\usepackage{booktabs}
\usepackage{tikz}
\usepackage{xcolor}
\usepackage{dirtytalk}  
\usetikzlibrary{positioning, calc, arrows.meta, shapes.misc}
\usepackage{cite}

\def \ie {i.e., }
\def \eg {e.g., }

\begin{document}
\title{TubeLite: Lightweight Multi-Actor Spatio-Temporal Action Detection}
%
%
\author{Ali Soltaninezhad\inst{1}\orcidID{0000-0002-1160-4112} \and
Melissa Cote\inst{1}\orcidID{0000-0002-5594-977X} \and
Alejandro Rico Espinosa\inst{1}\orcidID{0000-0002-4691-011X}
\and Tunai Porto Marques\inst{1}\orcidID{0000-0003-0850-9912} \and Alexandra Branzan Albu\inst{1}\orcidID{0000-0001-8991-0999}}
\authorrunning{A. Soltaninezhad et al.}
%
\institute{Electrical and Computer Engineering, 
University of Victoria, Victoria, BC, Canada\\
\email{\{alisoltaninezhad, mcote, arico, tunaip, aalbu\}@uvic.ca}}
\maketitle              
\begin{abstract}
Spatio-temporal action detection in videos requires jointly localizing actors in space and identifying action boundaries over time. 
A common challenge is constructing temporally stable action tubes, as frame-level detectors often suffer from jitter, fragmentation, and imprecise temporal localization. Many recent approaches address this by introducing heavy spatio-temporal transformers or optical-flow-based pipe\-lines, leading to high computational cost and limited scalability. We propose TubeLite, a lightweight framework for spatio-temporal action detection that focuses on stable tube construction and boundary-aware temporal modeling. TubeLite represents each actor as a tube, defined as a sequence of bounding boxes associated with a single actor over time, and explicitly enforces temporal consistency at both the spatial and semantic levels. The method combines low-jitter actor detection, Gaussian-weighted actor feature extraction, efficient short-term temporal propagation, and a boundary-focused temporal prediction head, while avoiding optical flow and large-scale temporal attention. Despite its compact design, TubeLite achieves strong video-level localization performance. It improves Video-mAP@0.5 by 4.5 and 7.1 percentage points over the best compared method on the MultiSports and UCF101-24 datasets, respectively, with substantially fewer parameters and floating-point operations than transformer-based alternatives, demonstrating that effective spatio-temporal action detection can be obtained through principled, lightweight temporal modeling.

\end{abstract}

\vspace{-8pt}
\section{Introduction}
\label{sec:intro}
\vspace{-4pt}

Spatio-temporal action detection (STAD) aims to localize actors and track their actions over time in videos. It is a fundamental capability of video understanding systems for dynamic and typically unconstrained scenes, deployed in various applications such as surveillance, sports analytics, and onboard vision for autonomous platforms. Despite substantial progress, STAD remains challenging because models must jointly handle (i) spatial detection under heavy occlusion, motion blur, and appearance variability, (ii) temporal reasoning over long and dynamically changing video sequences, and (iii) multi-actor interactions that shape the semantics of actions. These demands make the STAD problem distinct from both frame-level classification and short-range action recognition.

Many recent STAD architectures require heavy 3D backbones, optical flow computation, or transformer stacks with quadratic complexity, yielding high training costs and making it difficult to process long sequences. Moreover, existing frameworks typically rely on proposal-based pipelines or external tracking mechanisms, which introduce error cascades and reduce robustness in multi-actor scenes. In practical deployment scenarios, models must remain computationally efficient while still capturing the temporal structure necessary for precise action boundary identification. We observe that many errors in STAD pipelines arise not from insufficient visual capacity, but from the \emph{temporal fragmentation} introduced by frame-independent detectors, noisy actor trajectories, and insufficient long-range temporal smoothing. This motivates an architecture that (i) uses lightweight but reliable spatial detection, (ii) extracts temporally coherent actor representations, (iii) models temporal evolution without expensive 3D reasoning, and (iv) produces stable, temporally consistent actor tubes, \ie sequences of bounding boxes (BBs) associated with a single actor over time.

\vspace{-6pt}
\subsection{Contributions}
\label{sec:intro_cont}
\vspace{-4pt}

We propose TubeLite, a lean and efficient architecture for STAD that integrates reliable spatial detection, actor-centric feature extraction, and long-range temporal modeling within a single fully-differentiable framework (see Sec.~\ref{sec:runtime} for a quantitative efficiency analysis). Our contributions are summarized as follows: (i) \textbf{Lightweight spatial detector} built on ConvNeXt \cite{liu2022convnext} combined with a Center\-Net-inspired detection head \cite{zhou2019centernet}. This provides high-quality actor boxes without expensive 3D convolutions or region proposal networks. (ii) \textbf{Gaussian Region of Interest (ROI) tokenization with motion cues.} We propose a differentiable actor-centric embedding mechanism, extending A$^2$-Nets~\cite{sun2018actor} to incorporate feature-level motion cues using a Gaussian-weighted region extractor, producing stable actor tokens without reliance on optical flow, 3D CNNs, or heavy temporal attention. (iii) \textbf{Gated recurrent unit (GRU)-based temporal propagation for actor-centric smoothing.} The GRU \cite{cho2014gru} propagates information through time, stabilizing actor representations and reducing temporal jitter compared to pure attention-based models. (iv) \textbf{Temporal head for boundary-aware action detection.} We design a multi-branch temporal prediction head that outputs frame-level class scores, \say{actionness}, start/end boundaries, and distance-to-boundary regressions, drawing inspiration from temporal proposal networks \cite{lin2018bsn,lin2019bmn} while remaining actor-conditioned. (v) \textbf{Tube-aware temporal regularization.} We propose a tube-aware loss that promotes smooth, consistent actor boxes and action probabilities over time, enabling stable video-level detections without requiring extra labels or optical flow.

TubeLite achieves a strong video-level performance on the MultiSports \cite{li2021multisports} and UCF101-24 \cite{soomro2012ucf101} datasets, outperforming several heavier architectures while running significantly faster. We show that carefully integrating lightweight temporal reasoning and stabilizing losses can close much of the gap between light\-weight detectors and large STAD models.

\vspace{-7pt}
\section{Related Work} 
\label{sec:relworks}
\vspace{-5pt}

In this section, we first divide STAD approaches according to their number of stages. We then review the STAD literature according to the following topics: temporal stability, temporal boundary modeling, and lightweight architectures.

Early methods used a two-stage pipeline, \ie frame detection then linked into action tubes using tracking or post-hoc association \cite{gkioxari2015contextual,gu2018ava,he2017maskrcnn}, which was effective but noise-sensitive and slow due to offline processing. Newer single-stage models (\eg STCA \cite{tian2025stca}, SlowFast \cite{wu2019slowfast}, and transformer-based detectors such as TubeR \cite{Zhao2022tuber} and MeMViT \cite{wu2022memvit}) jointly handle spatial and temporal modeling via space-time convolutions or self-attention, achieving strong results on benchmarks such as UCF101-24 \cite{soomro2012ucf101}, AVA \cite{gu2018ava}, and MultiSports \cite{li2021multisports}. However, their heavy backbones, global attention, or long temporal windows hinder fast inference.

Maintaining stable predictions across frames in STAD is challenging due to detector jitter, occlusion, and crowded scenes. Several prior works addressed this indirectly via tube linking \cite{gu2018ava}, attention-based propagation \cite{wu2019slowfast}, or space–time aggregation in transformer layers \cite{Zhao2022tuber, wu2022memvit}. Others used tracking-style mechanisms or motion cues, adding overhead; this motivates simpler consistency methods without heavy global operators or tracking modules.

Accurate action start/end localization is critical in STAD, especially for short actions or densely annotated videos. Temporal action detectors like BSN \cite{lin2018bsn}, BMN \cite{lin2019bmn}, and GTAD \cite{xu2020gtad} improved this with boundary-sensitive objectives. Recent models like ActionFormer \cite{zhang2022actionformer} extend this to spatial–temporal settings using frame-level actionness, boundary logits, and distance estimates. Such boundary-aware designs enhance supervision for ambiguous transitions and improve robustness to partial occlusion and rapid motion.

High computational cost for long, high-resolution clips has motivated research on efficient designs. ConvNeXt \cite{liu2022convnext} showed modern CNNs can match early transformers at lower cost, while TSM \cite{lin2019tsm}, X3D \cite{feichtenhofer2020x3d}, and RTD-Net \cite{tan2021rtdnet} improved efficiency via channel shifts, depthwise factorization, or anchor-free detection. In STAD, lightweight models are crucial for handling many frames and actors per clip, motivating approaches that preserve temporal coherence without expensive global modules or long-range transformers.

\vspace{-6pt}
\section{Method}
\label{sec:method}
\vspace{-4pt}

\begin{figure}[t]
    \centering
    \includegraphics[width=0.90\textwidth]{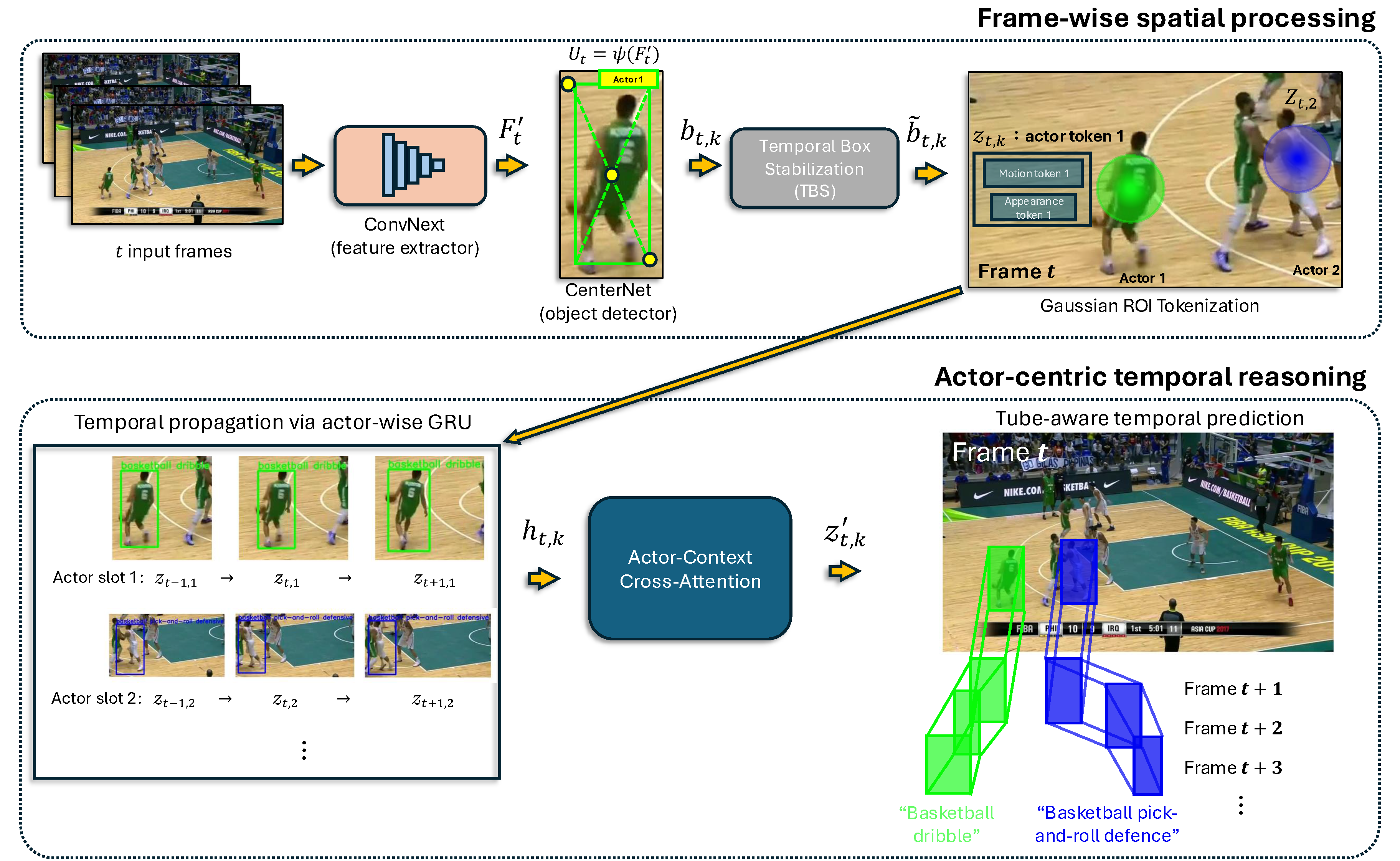}
    \vspace{-10pt}
    \caption{Overview of the TubeLite framework for STAD. 
    Top: frame-wise spatial processing, where short video clips are processed independently per frame using a 2D backbone and a CenterNet-style detector. Detected actor boxes are temporally stabilized (TBS) and converted into compact actor tokens via Gaussian ROI tokenization that fuses appearance and motion cues.
    Bottom: actor-centric temporal reasoning, where fixed actor slots are associated across frames, propagated through time using actor-wise GRU modules, and enriched with frame-level global context via actor-context cross-attention with learned latent tokens. The resulting actor representations are used for tube-aware prediction of action classes and temporal boundaries across the clip.}
    \label{fig:tubelite_overview}
    \vspace{-12pt}
\end{figure}

TubeLite is designed around the notion that STAD fundamentally depends on the construction of accurate and temporally stable actor tubes. Unlike recent transformer-heavy pipelines, TubeLite emphasizes actor-centric reasoning with minimal computational overhead. The architecture integrates: (i) low-jitter detection, (ii) Gaussian-weighted motion-aware tokenization, (iii) lightweight temporal and relational propagation, and (iv) a boundary-focused temporal classification module. Fig.~\ref{fig:tubelite_overview} provides an overview of the full architecture, illustrating the main processing steps: frame-wise spacial processing (top, Sec.~\ref{sec:method_framewise}) and actor-centric temporal reasoning (bottom, Sec.~\ref{sec:method_temporal}).

\textbf{Actor tube definition}: In STAD, an \emph{actor tube} is a temporally ordered sequence of spatial BBs corresponding to a single actor instance over a video segment.
In TubeLite, tubes are represented using fixed \emph{actor slots}, which serve as model-side proxies for actor instances within a short video clip.

Formally, for actor slot $k \in \{1,\dots,K\}$, a tube is represented as:
\vspace{-6pt}
\begin{equation}
\mathcal{T}_k = \{ b_{t,k} \}_{t=1}^{T}, \quad b_{t,k} \in [0,1]^4,
\vspace{-6pt}
\end{equation}
where $b_{t,k}$ denotes the normalized BB associated with slot $k$ at frame $t$, and $T$ is the number of frames in the input clip. A slot may correspond to a visible actor instance or remain inactive if fewer than $K$ actors are present. Each tube is associated with a temporal action label and boundary annotations indicating the start and end of the action within the clip. TubeLite constructs tubes over short, overlapping clips and refines them through temporal stabilization and recurrent propagation, rather than explicit long-term identity tracking.

\vspace{-6pt}
\subsection{Frame-wise Spatial Processing}
\label{sec:method_framewise}
\vspace{-4pt}

\textbf{Feature extraction and object detection:} TubeLite processes video clips by extracting spatial features and actor candidates independently at each frame, deferring temporal reasoning to later actor-centric modules. Given an input clip
$X \in \mathbb{R}^{T \times 3 \times H \times W}$, where $H$ and $W$ are the height and width of a frame, a shared 2D convolutional backbone ($\phi$) is applied frame-wise to obtain feature maps:
\vspace{-5pt}
\begin{equation}
F_t = \phi(X_t) \in \mathbb{R}^{C \times H_f \times W_f},
\vspace{-5pt}
\end{equation}
followed by a $1{\times}1$ projection ($W_{\mathrm{proj}}$):
\vspace{-5pt}
\begin{equation}
F_t' = W_{\mathrm{proj}} F_t \in \mathbb{R}^{D \times H_f \times W_f},
\vspace{-7pt}
\end{equation}
which aligns feature dimensionality across all subsequent modules, where D denotes the shared embedding dimension used throughout TubeLite. This stage performs no temporal feature mixing, ensuring that early processing remains purely spatial and computationally lightweight. Actor candidates are localized from $F_t'$ using a modified CenterNet-style
detector that predicts actor center heatmaps and BB geometry. Unlike the original formulation, TubeLite replaces the independent prediction heads with a \emph{shared convolutional tower} that produces a task-consistent
intermediate representation ($U_t$):
\vspace{-7pt}
\begin{equation}
U_t = \psi(F_t'),
\vspace{-7pt}
\end{equation}
where $\psi(\cdot)$ denotes the shared convolutional tower applied to projected frame-wise features. Shared towers have been explored in image-based detectors such as RetinaNet~\cite{lin2017focal}, but their role is particularly important in STAD, where inconsistencies in frame-level predictions can accumulate over time and lead to unstable actor tubes. Forcing heatmap estimation and box regression to rely on a common representation reduces frame-to-frame jitter and improves the temporal reliability of detected actor locations. Concretely, $U_t$ feeds a heatmap head and a regression head; the top-$K$ detections yield raw boxes $b_{t,k}$ that populate the actor tubes $\mathcal{T}_k$. Rather than maximizing per-frame recall, the detector retains a fixed set of the $K$ most temporally reliable actor candidates at each frame. These candidates serve as stable spatial anchors for subsequent tube construction, temporal stabilization, and actor-centric reasoning.

\textbf{Temporal box stabilization (TBS)}: Per–frame detectors in videos often exhibit minor spatial fluctuations due to motion blur, partial occlusion, and heatmap noise.  
To obtain temporally consistent actor trajectories before tokenization, we introduce a lightweight TBS step. TBS applies a standard linear temporal filter widely used in multi-object tracking~\cite{bewley2016simple}, adopted here as a deliberate lightweight design choice that requires no learned parameters. For each actor slot $k$ at time $t$, a stabilized box $\tilde{b}_{t,k}$ is derived from the raw prediction $b_{t,k}$ as:
\vspace{-7pt}
\begin{equation}
\tilde{b}_{t,k} = \lambda\, b_{t,k} + (1-\lambda)\,\frac{b_{t-1,k}+b_{t+1,k}}{2}.
\vspace{-7pt}
\end{equation}
The term $b_{t,k}$ denotes the detector output at frame $t$ (normalized $(x_1,y_1,x_2,y_2)$ coordinates), while
$b_{t-1,k}$ and $b_{t+1,k}$ provide short-range temporal support, leveraging local temporal context without requiring additional learned parameters. The averaging term models a first-order motion prior, and the scalar $\lambda \in [0,1]$ balances responsiveness to instantaneous detections against temporal smoothness. The resulting $\tilde{b}_{t,k}$ reduces high-frequency jitter, mitigates isolated dropouts, and provides a more stable spatial reference for the subsequent stage. The stabilized boxes $\tilde{b}_{t,k}$ replace $b_{t,k}$ in tube $\mathcal{T}_k$ and define the Gaussian mask center $c_{t,k}$ used to extract actor tokens $z_{t,k}$.

\textbf{Gaussian ROI tokenization}: This stage extracts compact, motion-sensi\-tive actor descriptors. The design is inspired by Gaussian-based attention in A$^2$-Nets~\cite{sun2018actor}, and extended to incorporate feature-level motion cues.

Given a stabilized box $\tilde{b}_{t,k}$, we define a differentiable Gaussian mask over the feature map as:
\vspace{-7pt}
\begin{equation}
G_{t,k}(x,y)
\propto
\exp\!\left(
    -\frac{\|(x,y) - c_{t,k}\|^2}{2\sigma_g^2}
\right),
\vspace{-5pt}
\end{equation}
where $(x,y)$ indexes spatial locations on the feature map, $c_{t,k} \in \mathbb{R}^2$ denotes the center of $\tilde{b}_{t,k}$ expressed in feature-map coordinates, and
$\sigma_g$ controls the spatial extent of the Gaussian weighting. The proportionality symbol indicates that $G_{t,k}$ is normalized such that $\sum_{x,y} G_{t,k}(x,y) = 1$.

The appearance descriptor for actor slot $k$ at time $t$ (appearance token $a_{t,k} \in \mathbb{R}^{D}$) is computed by Gaussian-weighted pooling of the projected feature map:
\vspace{-7pt}
\begin{equation}
a_{t,k}
= \sum_{x,y} G_{t,k}(x,y)\, F_t'(:,x,y),
\vspace{-9pt}
\end{equation}
where $F_t'(:,x,y) \in \mathbb{R}^{D}$ denotes the feature vector at spatial location $(x,y)$. 

Short-term motion information is captured via residual features:
\vspace{-7pt}
\begin{equation}
\Delta F_t' = F_t' - F_{t-1}',
\vspace{-7pt}
\end{equation}
where $\Delta F_t' \in \mathbb{R}^{D \times H_f \times W_f}$ encodes temporal changes in feature activations between consecutive frames. These residual features are aggregated using the same Gaussian mask:
\vspace{-7pt}
\begin{equation}
m_{t,k}
= \sum_{x,y} G_{t,k}(x,y)\, \Delta F_t'(:,x,y),
\vspace{-7pt}
\end{equation}
yielding a motion-sensitive token $m_{t,k} \in \mathbb{R}^{D}$ (without explicit optical flow).

Appearance and motion tokens are adaptively fused through a lightweight gating mechanism:
\vspace{-7pt}
\begin{equation}
\alpha_{t,k}
= \mathrm{sigmoid}\!\left(W [a_{t,k}; m_{t,k}]\right),
\vspace{-3pt}
\end{equation}
where $[a_{t,k}; m_{t,k}] \in \mathbb{R}^{2D}$ denotes channel-wise concatenation,
$W \in \mathbb{R}^{1 \times 2D}$ is a learned linear projection, and
$\alpha_{t,k} \in (0,1)$ is a scalar gate controlling the relative contribution of appearance and motion.

The final actor token $z_{t,k} \in \mathbb{R}^{D}$ is computed as:
\vspace{-8pt}
\begin{equation}
z_{t,k}
= \alpha_{t,k}\, a_{t,k}
+ (1-\alpha_{t,k})\, m_{t,k},
\vspace{-8pt}
\end{equation}
producing a motion-aware descriptor that preserves spatial localization through Gaussian pooling while adapting to temporal change.

\vspace{-8pt}
\subsection{Actor-centric Temporal Reasoning}
\label{sec:method_temporal}
\vspace{-6pt}

\textbf{Temporal propagation via actor-wise gated recurrent unit (GRU)}: To enrich per-frame actor descriptors with temporal context, TubeLite applies a GRU~\cite{cho2014gru} independently to each actor slot. Given a sequence of slot-wise tokens $\{z_{t,k}\}_{t=1}^T$, the GRU evolves a latent temporal state according to:
\vspace{-5pt}
\begin{equation}
h_{t,k} = \mathrm{GRU}\big(z_{t,k},\, h_{t-1,k}\big),
\vspace{-7pt}
\label{eq:GRU}
\end{equation}
where the recurrent update integrates the current actor token with its historical state, producing a temporally contextualized representation $h_{t,k}$. This recurrent formulation models the temporal progression of each actor, such as gradual pose changes, repetitive motion patterns, and action onsets, through an actor-wise state evolution. Temporal propagation is performed independently for each actor slot, enforcing a strong inductive bias that prioritizes intra-actor temporal coherence over cross-actor mixing at this stage. This separation reduces temporal noise caused by nearby or transient actors and stabilizes tube evolution in crowded scenes, where early fusion of actor dynamics can lead to identity drift. Contextual and multi-actor cues are incorporated at later stages of the model, allowing TubeLite to retain lightweight temporal propagation while still capturing global structure when needed. In contrast to clip-level transformers such as TimeSformer~\cite{bertasius2021timesformer}, which apply full space-time self-attention, the actor-wise GRU introduces a computationally lightweight temporal model with several advantages: (i) lower latency due to linear-time recurrence rather than quadratic attention, (ii) significantly fewer parameters while retaining expressive temporal memory, and (iii) an inherent bias toward temporal smoothness, which suppresses frame-level noise and stabilizes evolving actor representations. The resulting sequence $\{h_{t,k}\}$ provides temporally enriched actor representations that are directly consumed by the subsequent interaction and prediction modules.

\begin{figure}[t]
    \centering
    \includegraphics[width=0.78\textwidth]{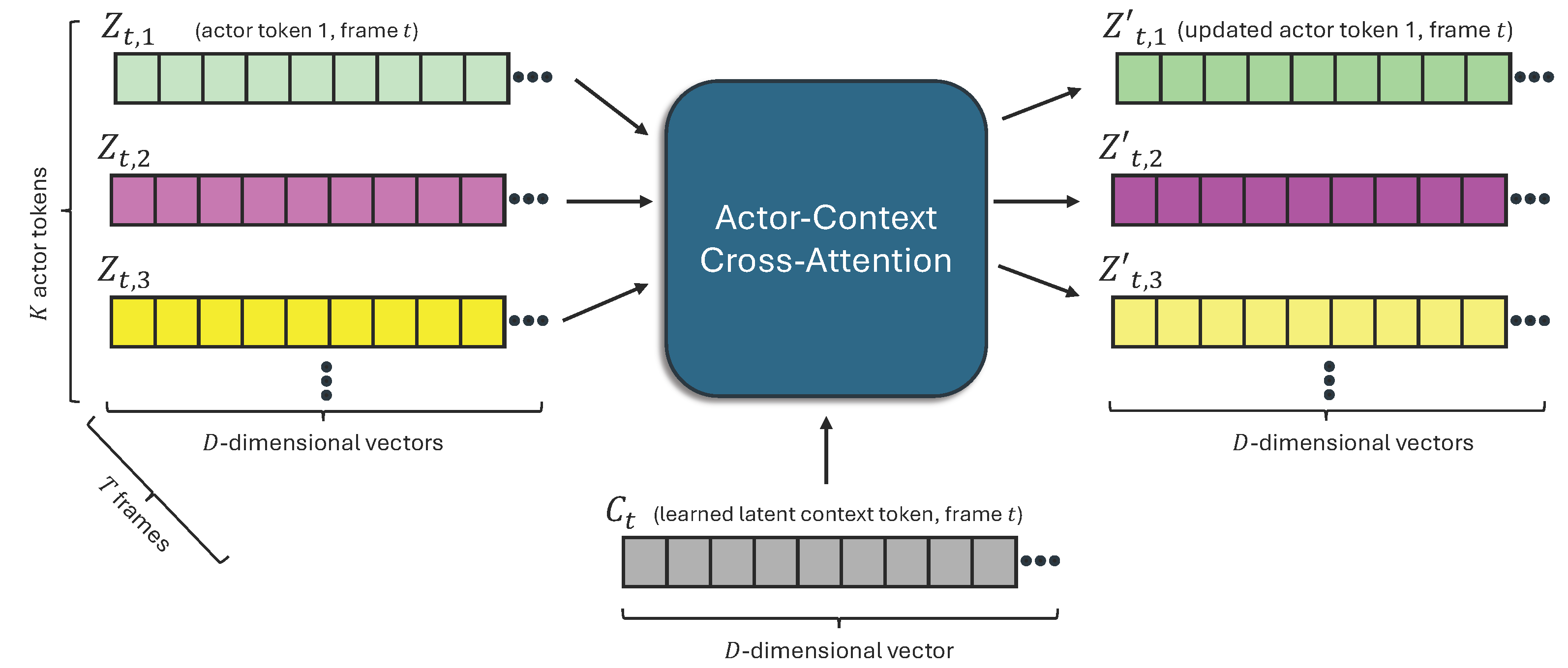}
    \vspace{-12pt}
    \caption{Actor-context cross-attention with latent context tokens. Cross-attention at a single frame \(t\), where a fixed set of \(K\) actor tokens \(z_{t,k} \in \mathbb{R}^{D}\) interacts with a learned latent context token \(C_t \in \mathbb{R}^{D}\). $C_t$ is a learned parameter (not derived from image features) that provides a shared frame-level summary and modulates actor representations. This operation is applied independently across the \(T\) frames of the input clip.}
    \label{fig:actorcontext}
    \vspace{-16pt}
\end{figure}

\textbf{Actor-context cross-attention}: To incorporate structured global context without the cost of full actor-actor or spatiotemporal self-attention, TubeLite introduces a lightweight \emph{actor-context cross-attention} mechanism based on learned latent context tokens (Fig.~\ref{fig:actorcontext}). The design is inspired by cross-attention pooling in Vision Transformers~\cite{dosovitskiy2020vit} and the latent bottleneck formulation used in Perceiver-style architectures~\cite{jaegle2021perceiver}, but is applied strictly within each frame.

For an input clip of $T$ frames, we associate each frame $t$ with a learned latent context token $C_t \in \mathbb{R}^{1 \times D}$ and collect them as $C \in \mathbb{R}^{T \times 1 \times D}$, with $D$ being the shared embedding dimension. These context tokens are learned parameters that do not correspond to image patches or pixels; instead, they serve as compact latent variables that summarize the configuration of actors present in each frame. Each actor representation (here a D-dimensional token associated with actor slot $k$ at frame $t$) is given by $z_{t,k} \in \mathbb{R}^{T \times K \times D}$. A cross-attention block is applied independently at each frame:
\vspace{-8pt}
\begin{equation}
(z'_t,\, C'_t) = \mathrm{XAttn}(z_t,\, C_t),
\vspace{-6pt}
\end{equation}
where $\mathrm{XAttn}(\cdot)$ denotes a multi-head cross-attention operator. The context token $C_t$ attends over the set of actor tokens $\{z_{t,k}\}_{k=1}^{K}$ at the same frame, producing an updated context representation $C'_t$, while actor tokens are simultaneously modulated by feedback from $C_t$ to yield updated actor representations $z'_t$.

The attention weights are computed from similarity scores between the context token and actor embeddings, allowing $C_t$ to aggregate a global summary of the actor configuration present at frame $t$. Conversely, actor tokens receive contextual information that reflects the overall scene configuration, improving robustness in crowded or
ambiguous settings. This design provides three advantages. First, it injects a shared frame-level context signal that anchors actor representations without explicit actor-actor attention. Second, it regularizes actor tokens by allowing them to reference a common latent context rather than relying solely on localized evidence. Third, it enables efficient aggregation: attention is evaluated only between $K$ actor tokens and a single latent token per frame, avoiding the quadratic cost of full self-attention. Importantly, this module performs no temporal attention: context tokens are frame-indexed and do not propagate across time. Temporal dependencies are modeled exclusively by the actor-wise GRU (see Eq.~\ref{eq:GRU}). The updated actor tokens $z'_t$ are passed to the temporal prediction head, while the updated context tokens $C'_t$ serve as auxiliary latent summaries used for boundary and action prediction.

\textbf{Tube-aware temporal prediction}: For each actor slot $k \in \{1,\dots,K\}$ at frame $t \in \{1,\dots,T\}$, the temporally enriched representation $h_{t,k} \in \mathbb{R}^{D}$ is mapped to a set of frame-level temporal predictions that jointly describe \emph{what} action is occurring and \emph{where} it begins and ends in time. Specifically, the temporal head produces, for each $(t,k)$ pair, class logits over the action label set, an actionness score indicating whether an action is present, start
and end boundary logits, left and right temporal distance predictions, and a center-quality score reflecting proximity to the temporal center of the action.

We adopt boundary-sensitive supervision strategies inspired by BSN~\cite{lin2018bsn}, BMN~\cite{lin2019bmn}, and TSP~\cite{alwassel2021tsp}, and adapt them to an actor-centric, tube-based formulation. For each GT actor tube, its temporal extent is annotated by a start frame index $s$ and an end frame index $e$, with $1 \leq s \leq e \leq T$. From these annotations, we derive dense frame-level supervisory signals for each actor--frame pair $(t,k)$, including: (i) an \emph{actionness} indicator specifying whether frame $t$ lies inside the action interval $[s,e]$; (ii) \emph{start} and \emph{end} boundary targets encoding proximity to $s$ and $e$; (iii) normalized \emph{left} and
\emph{right} temporal distances to the boundaries; and (iv) a \emph{center-quality} score that peaks at the temporal midpoint $(s+e)/2$ and decays toward the boundaries. Together, these signals encourage precise temporal localization rather than coarse segment classification.

In addition to per-frame supervision, we enforce holistic alignment between predicted and GT action segments using a temporal Intersection-over-Union (tIoU) loss. For each clip, predicted start and end boundaries $\hat{s}$ and $\hat{e}$ are obtained by aggregating boundary logits across the $K$ actor slots at each frame, while $s$ and $e$ denote the GT temporal boundaries. The tIoU loss is defined as:
\vspace{-7pt}
\begin{equation}
\mathcal{L}_{\mathrm{tIoU}} =
1 - \mathrm{tIoU}(\hat{s}, \hat{e}, s, e),
\vspace{-7pt}
\end{equation}
where $\mathrm{tIoU}(\cdot)$ measures the temporal overlap between the predicted interval $[\hat{s},\hat{e}]$ and the GT interval $[s,e]$. This objective directly penalizes segment-level misalignment and promotes coherent tube-level boundary estimation.

To further promote temporal stability in both localization and semantics, we introduce a lightweight tube-aware regularization. Let $\hat{b}_{t,k} \in [0,1]^4$ denote the predicted BB for actor slot $k$ at frame $t$, represented by normalized $(x_1,y_1,x_2,y_2)$ coordinates, and let $\hat{p}_{t,k} \in [0,1]$ denote the corresponding predicted actionness score. Temporal smoothness is encouraged via:
\vspace{-8pt}
\begin{equation}
\mathcal{L}_{\mathrm{tube}}
=
\frac{1}{T-1}
\sum_{k=1}^{K}
\sum_{t=1}^{T-1}
\left(
\left\|\hat{b}_{t+1,k} - \hat{b}_{t,k}\right\|_1
\;+\;
\lambda_{p}\,\left|\hat{p}_{t+1,k} - \hat{p}_{t,k}\right|
\right),
\label{eq:tube_loss}
\vspace{-8pt}
\end{equation}
where $\|\cdot\|_1$ denotes the $\ell_1$ norm and $\lambda_{p}$ balances spatial and semantic smoothness. This regularizer reduces jitter, stabilizes tube geometry, and yields more consistent temporal action boundaries.

All components of TubeLite are trained jointly in an end-to-end manner using a weighted multi-term objective that integrates spatial detection, set-based actor supervision, temporal boundary modeling, and tube-level regularization:
\vspace{-8pt}
\begin{equation}
\mathcal{L}
=
\alpha_1\mathcal{L}_{\mathrm{det}}
+
\alpha_2\mathcal{L}_{\mathrm{set}}
+
\alpha_3\mathcal{L}_{\mathrm{temp}}
+
\alpha_4\mathcal{L}_{\mathrm{tube}},
\label{eq:loss}
\vspace{-8pt}
\end{equation}
where $\alpha_i$ are weights, $\mathcal{L}_{\mathrm{det}}$ denotes the CenterNet-style detection losses, including focal loss for center heatmap prediction and regression losses for BB size and offsets~\cite{zhou2019centernet,lin2017focal},
$\mathcal{L}_{\mathrm{set}}$ comprises the set-based actor supervision induced by Hungarian matching~\cite{carion2020detr} and Sinkhorn-based soft assignment \cite{cuturi2013sinkhorn}, $\mathcal{L}_{\mathrm{temp}}$ aggregates the boundary-aware temporal losses (actionness, start/end boundaries, temporal distance regression, center quality, and tIoU), inspired by prior boundary-sensitive temporal localization
methods~\cite{lin2018bsn,lin2019bmn,alwassel2021tsp}, and $\mathcal{L}_{\mathrm{tube}}$ is the tube-aware temporal smoothing term defined in Eq.~\eqref{eq:tube_loss}. All loss terms are computed on the same forward pass and optimized jointly, with fixed weights (see Sec.~\ref{sec:training_details}) chosen to balance spatial accuracy, temporal localization precision, and tube stability.

\vspace{-6pt}
\section{Experiments}
\label{sec:exp}
\vspace{-4pt}

This section presents the experimental evaluation of TubeLite on two established benchmarks for STAD: UCF101-24 \cite{soomro2012ucf101} and MultiSports \cite{li2021multisports}. It describes the evaluation protocol and the datasets, gives implementation details, provides a quantitative comparison with the state of the art, a qualitative evaluation, a runtime analysis, and ends with an ablation study.

\vspace{-8pt}
\subsection{Evaluation Protocol}
\vspace{-6pt}

We adopt the standard STAD evaluation metrics used in prior work \cite{Zhao2022tuber,tian2025stca,gu2018ava}.

\textbf{Frame-level evaluation}: Frame-mAP@0.5 measures spatial localization accuracy independently at each frame. This metric primarily reflects the quality of the detection head, Gaussian ROI token extraction, and short-term spatial consistency produced by the temporal propagation module.

\textbf{Video-level evaluation}: 
Video-mAP@0.5 evaluates full spatio-temporal tubes and is therefore sensitive to temporal segmentation quality. Following TubeR \cite{Zhao2022tuber}, predicted tubes are matched to GT tubes using spatio-temporal IoU. This metric particularly highlights the contribution of TubeLite's temporal boundary estimation (start, end, distances) and tube-aware regularization, which discourage fragmented or inconsistent predictions.

\vspace{-10pt}
\subsection{Datasets}
\vspace{-4pt}

UCF101-24 \cite{soomro2012ucf101} contains 24 action categories annotated with spatio-temporal tubes across approximately 3,200 videos. Following common practice \cite{Zhao2022tuber,he2017maskrcnn}, we use split~1 for training and evaluation. The dataset is characterized by rapid camera motion, significant scale variation, and temporally fragmented actions, making precise temporal boundary estimation challenging. MultiSports \cite{li2021multisports} provides 3,200 densely annotated multi-actor videos spanning 66 action categories. The presence of multiple interacting actors per frame and fast motion dynamics makes it a demanding STAD benchmark. We follow the official split. Compared to UCF101-24, MultiSports places greater emphasis on robust actor association and temporal coherence across highly dynamic scenes.

\vspace{-10pt}
\subsection{Implementation Details}
\label{sec:training_details}
\vspace{-4pt}

\textbf{Input processing}: Videos are processed as clips of $T=16$ $224 \times 224$ frames with a temporal stride of~8. Motion cues derive from feature-level frame differences.

\textbf{Backbone and detection}: Spatial features are extracted using a Conv\-NeXt-Tiny backbone~\cite{liu2022convnext} applied independently to each frame. A CenterNet-style detection head~\cite{zhou2019centernet} predicts actor center heatmaps, BB sizes, and offsets from the projected feature maps. Following standard CenterNet practice, the heatmap head is initialized with a negative bias to suppress early false positives, and BB dimensions are regressed in log-space to improve numerical stability. We retain the top-$K=5$ actor candidates per frame. This value corresponds to the maximum number of annotated actors in both UCF101-24 and MultiSports, ensuring full coverage without introducing redundant actor slots.

\textbf{Set-based actor assignment}: Since TubeLite predicts a fixed, unordered set of actor slots per frame, we employ set-based supervision following DETR-style bipartite matching~\cite{carion2020detr}. For each frame, predicted actor slots are matched to GT actors using the Hungarian algorithm based on a weighted combination of classification confidence, bounding-box distance, and generalized IoU~\cite{rezatofighi2019giou}. Only matched slot–actor pairs contribute to classification and localization losses;
unmatched slots receive no box or class supervision. We also compute a differentiable soft assignment using Sinkhorn normalization~\cite{cuturi2013sinkhorn}, yielding graded correspondences between slots and GT actors. A lightweight temporal regularization encourages consistency of these soft assignments across adjacent frames, reducing identity swit\-ches without explicit tracking. Actor-slot presence is supervised using a binary classification loss derived from the matching results.

\textbf{Optimization and loss weighting}: All components of TubeLite are trained end-to-end using the AdamW optimizer~\cite{loshchilov2019decoupled} with a learning rate of $1\times10^{-4}$ and weight decay of $1\times10^{-4}$. Automatic mixed-precision training is used to reduce memory usage and improve training throughput. Regarding the loss (Eq.~\ref{eq:loss}), all weights are fixed and shared across datasets; detection and classification terms are emphasized to ensure stable actor localization and slot assignment ($\alpha_1$ and $\alpha_2$ = 1), while temporal boundary and tube-smoothing losses are weighted more conservatively ($\alpha_3$ and $\alpha_4$ = 0.1) to refine temporal structure without destabilizing spatial predictions.

\vspace{-8pt}
\subsection{Quantitative Evaluation}
\vspace{-6pt}

We compare TubeLite's performance against a broad family of STAD approa\-ches, including anchor-free RGB detectors (ROAD~\cite{singh2017road}, YOWO~\cite{kopuklu2019yowo}),
multi-object tube trackers (MOC~\cite{li2020moc}), transformer-based pipelines (TubeR~\cite{Zhao2022tuber}), and two-stream architectures that leverage optical flow or heavy temporal modeling (2in1~\cite{zhao2019dance}, STEP~\cite{yang2019step}, CFAD \cite{li2020cfad}). These methods represent the dominant design philosophies for STAD, ranging from proposal-driven systems to deep transformer encoders with large computational footprints. TubeLite, by contrast, uses a single RGB stream and a lightweight actor-centric temporal model, avoiding 3D convolutions, optical flow, and global self-attention. All compared results are taken from the original publications. Comparisons are conservative, as several methods use longer clip inputs or heavier backbones. For example, TubeR [41] uses 32-frame clips with a CSN-152 backbone, and two-stream methods such as CFAD [21] additionally use optical flow; TubeLite uses 16-frame RGB-only clips with a ConvNeXt-Tiny backbone.

\textbf{Results on MultiSports}: Table~\ref{tab:ms_sorted} reports results on MultiSports \cite{li2021multisports}. TubeLite obtains a frame-mAP@0.5 of 20.07, outperforming all prior RGB-only detectors and approaching the accuracy of significantly heavier SlowFast-based systems. More importantly, TubeLite achieves a video-mAP@0.5 of 14.20, a substantial improvement over all compared methods, suggesting that its temporal stabilization, actorwise propagation, and boundary-aware head jointly reduce tube fragmentation, a persistent challenge on this benchmark. Unlike prior detectors that rely on multi-branch temporal aggregation or optical flow, TubeLite leverages stable Gaussian-based tokenization and actorwise temporal propagation to maintain coherent motion patterns.

\begin{table}[t]
\scriptsize
\centering
\begin{tabular}{lcccc}
\toprule
Method & Stream & Input Resolution & V@0.5$\uparrow$ & F@0.5$\uparrow$ \\
\midrule
ROAD~\cite{singh2017road}      & RGB  & $300\times300$     & 0.00 & 3.90 \\
MOC (K=11)~\cite{li2020moc}    & RGB  & $288\times288$     & 0.62 & 25.22 \\
MOC (K=7)~\cite{li2020moc}     & RGB  & $288\times288$     & 0.77 & 22.51 \\
YOWO~\cite{kopuklu2019yowo}    & Two-stream & $224\times224$ & 0.87 & 9.28 \\
SlowOnly Det.~\cite{wu2019slowfast} & RGB & short side 256 & 5.50 & 16.70 \\
SlowFast Det.~\cite{wu2019slowfast} & Two-stream & short side 256 & 9.65 & \textbf{27.72} \\
\midrule
TubeLite (ours)     & RGB  & $224\times224$ & \textbf{14.20} & 20.07 \\
\bottomrule
\end{tabular}
\caption{STAD task performance evaluation on MultiSports \cite{li2021multisports}. Best in bold font.}
\label{tab:ms_sorted}
\vspace{-24pt}
\end{table}

\textbf{Results on UCF101-24}: Table~\ref{tab:ucf_sorted} reports results on UCF101-24 \cite{soomro2012ucf101}. Despite using a lightweight ConvNeXt-Tiny backbone, TubeLite attains a frame-mAP@0.5 of 82.1 (second best), competitive with I3D- and CSN-based models that rely on substantially larger backbones and, in many cases, auxiliary flow streams. It provides the strongest video-level localization: a video-mAP@0.5 of 71.7, outperforming all prior RGB-stream and two-stream models by at least 7 p.p., including flow-augmented methods such as CFAD. This highlights the effectiveness of TubeLite's boundary-aware temporal head and tube-aware regularization, which jointly improve temporal precision and reduce tube fragmentation, a key weakness of proposal-based and transformer-based STAD pipelines. 

\begin{table}[t]
\scriptsize
\centering
\begin{tabular}{lcccc}
\toprule
Method & Stream & Backbone & V@0.5 & F@0.5 \\
\midrule
T-CNN~\cite{hou2017tcaction}           & RGB  & C3D     & --   & 41.4 \\
TacNet~\cite{song2019tacnet}           & Two-stream & VGG   & 52.9 & 72.1 \\
ACT~\cite{kalogeiton2017action}         & Two-stream & VGG   & 51.4 & 67.1 \\
2in1~\cite{zhao2019dance}               & Two-stream & VGG   & 50.3 & 78.5 \\
MOC~\cite{li2020moc}                   & RGB  & DLA34   & 50.7 & 72.1 \\
STEP~\cite{yang2019step}               & Two-stream & I3D  & --   & 75.0 \\
I3D Det.~\cite{carreira2017i3d}        & Two-stream & I3D & 59.9 & 76.3 \\
TubeR~\cite{Zhao2022tuber}            & RGB  & CSN-152 & 58.4 & \textbf{83.2} \\
CFAD~\cite{li2020cfad}              & Two-stream & I3D & 64.6 & 72.5 \\
\midrule
TubeLite (ours)              & RGB  & ConvNeXt-T & \textbf{71.7} & 82.1 \\
\bottomrule
\end{tabular}
\caption{STAD task performance evaluation on UCF101-24 \cite{soomro2012ucf101}. Best in bold font.}
\label{tab:ucf_sorted}
\vspace{-12pt}
\end{table}

\begin{figure}[t]
    \centering
    \includegraphics[width=0.78\textwidth]{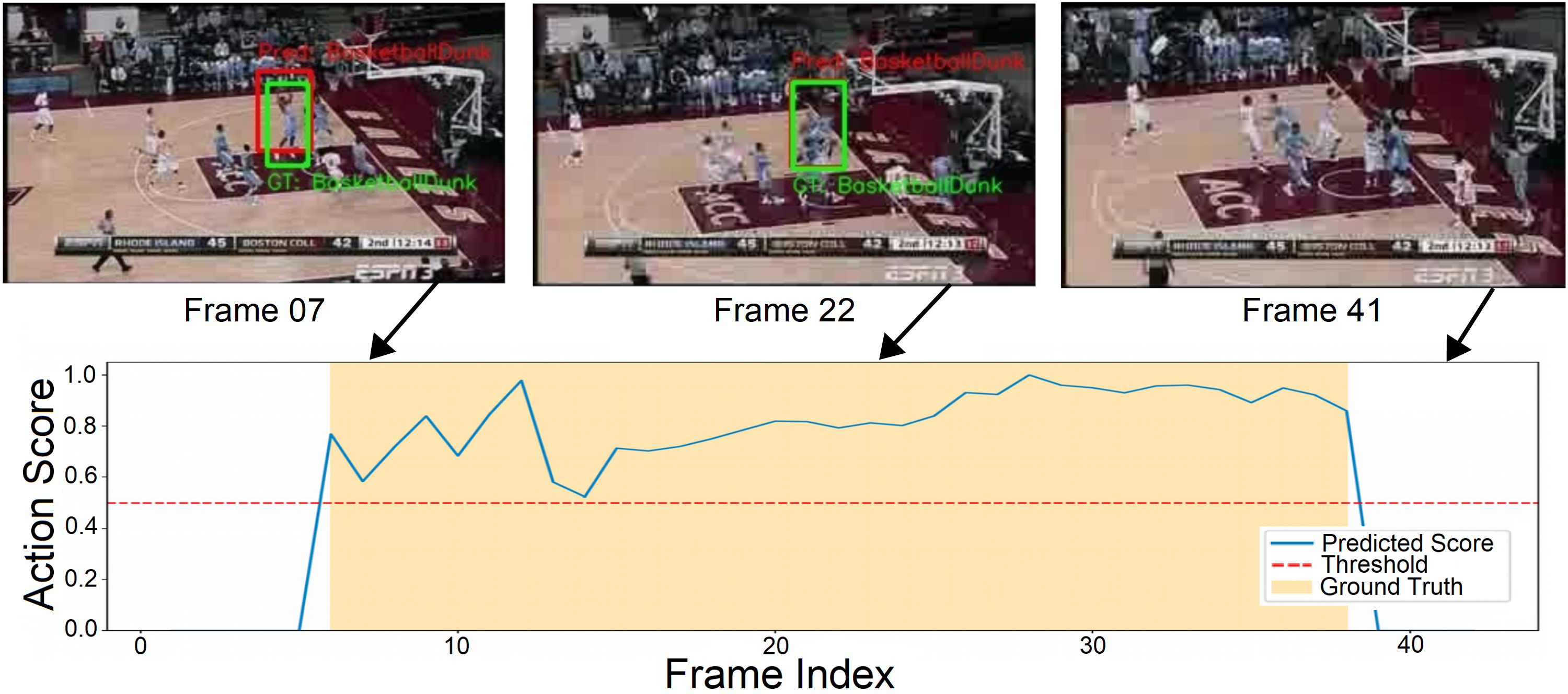}
    \vspace{-12pt}
    \caption{Sample STAD results on UCF101-24 (BasketballDunk action). Top: predicted actor BBs (red) match GT boxes (green) well. Bottom: prediction timeline (blue) above decision threshold (dashed line) during GT action interval (orange).}
    \label{fig:qualitative}
    \vspace{-12pt}
\end{figure}

\vspace{-10pt}
\subsection{Qualitative Evaluation}
\vspace{-8pt}

Fig.~\ref{fig:qualitative} shows representative TubeLite predictions on UCF101-24. The model produces temporally coherent tubes with stable spatial localization and well-aligned temporal boundaries. Predicted action scores remain above the decision threshold within the GT segment and rapidly decay outside the action interval.

\vspace{-10pt}
\subsection{Runtime Analysis}
\label{sec:runtime}
\vspace{-6pt}

We evaluate the inference efficiency of TubeLite using 16-frame input clips. Experiments are conducted on an NVIDIA H100 GPU for a fair comparison with recent spatio-temporal action detection systems. On average, TubeLite processes a 16-frame clip in 0.058~s, corresponding to 17.10~clips/s or 273.6~frames/s. It requires 61.96~GFLOPs per forward pass and contains 15.65M parameters. Compared to transformer-heavy STAD frameworks, TubeLite is substantially more computationally efficient. For reference, TubeR~\cite{Zhao2022tuber} reports inference at 156~frames/s, requires approximately 240~GFLOPs per clip, and has ~60M parameters. Other representative STAD systems, including CFAD~\cite{li2020cfad}, operate at around 130~frames/s with over 40M parameters, while many existing state-of-the-art methods report throughput in the range of 40--53~frames/s. In terms of backbone complexity, commonly used video models such as Slow\-Fast-50\cite{wu2019slowfast}, X3D-XL\cite{feichtenhofer2020x3d}, and CSN-152\cite{tran2019csn} require approximately 308, 290, and 342~GFLOPs per clip, respectively, about 5x more than TubeLite.

\vspace{-10pt}
\subsection{Ablation Study}
\vspace{-6pt}

We conduct an ablation study on UCF101-24 to assess the contribution of TubeLite’s core architectural components. Each variant removes a single module while keeping all other settings, training schedules, and evaluation protocols fixed. Performance is reported in Table~\ref{tab:ablation} using video-mAP@0.5 (V@0.5) and frame-mAP@0.5 (F@0.5). Removing the actor-wise GRU results in a clear drop in both video- and frame-level performance, confirming the importance of explicit temporal propagation for stabilizing actor representations and modeling action evolution over time. Eliminating the actor–context cross-attention leads to the largest degradation, particularly in video-mAP, highlighting the role of frame-level latent context in disambiguating actor behavior and improving robustness in multi-actor scenes without relying on expensive actor–actor attention. In contrast, removing tube-aware temporal supervision has little effect on frame-mAP but reduces video-mAP, indicating that boundary-focused and tube-level losses primarily enhance temporal coherence and segment-level localization rather than per-frame classification accuracy. Overall, this demonstrates that TubeLite’s performance gains arise from the complementary interaction of lightweight temporal propagation, structured actor–context reasoning, and tube-aware supervision.

\begin{table}[t]
\scriptsize
\centering
\label{tab:ablation}
\begin{tabular}{lcc}
\toprule
\textbf{Model Variant} & \textbf{V@0.5} & \textbf{F@0.5} \\
\midrule
TubeLite (Proposed) & \textbf{71.7} & \textbf{82.1} \\
\quad w/o Actor-Wise GRU & 69.3 & 79.7 \\
\quad w/o Actor-Context Cross-Attention & 67.5 & 78.7 \\
\quad w/o Temporal Losses (no $\mathcal{L}_{\mathrm{temp}}$) & 69.1 & 82.0 \\
\bottomrule
\end{tabular}
\caption{Ablation study removing TubeLite's components one at a time (UFC101-24).}
\vspace{-24pt}
\end{table}

\vspace{-8pt}
\section{Discussion and Conclusion}
\vspace{-6pt}

\textbf{Limitations}: TubeLite employs a clip-based inference strategy, processing short fixed-length windows (\eg 16 frames) rather than operating strictly online. As a result, predictions at a given time step may leverage limited future context within the clip (\eg for temporal stabilization and boundary estimation), which precludes true frame-by-frame deployment without look-ahead. Extending TubeLite to a strictly online setting would require modifying these temporal components to rely solely on past observations. In addition, the fixed actor-slot mechanism simplifies association and ensures stable temporal modeling, but assumes a bounded number of actors per clip. Although appropriate for datasets such as UCF101-24 and MultiSports, scenarios with highly variable crowd density could benefit from a dynamic or memory-driven slot allocation strategy. TubeLite also relies on lightweight temporal propagation rather than heavy full-sequence self-attention or 3D convolution, which contributes significantly to computational efficiency; however, more expressive long-range temporal modeling could further benefit performance on activities involving extended temporal context. 

\textbf{Potential negative impact}: As with all STAD systems, responsible and ethical deployment is essential, especially in settings involving surveillance or sensitive human activity monitoring.

\textbf{Conclusion}: Despite these constraints, TubeLite demonstrates that accurate and temporally stable action tubes can be produced with a compact architecture that avoids optical flow, heavy spatiotemporal attention, or large 3D backbones. By combining Gaussian ROI tokenization, actor-wise GRU propagation, clip-level cross-attention, and boundary-aware temporal supervision, TubeLite achieves strong video-mAP on UCF101-24 and MultiSports while maintaining a small computational footprint. The results highlight a promising direction for efficient STAD: structured, localized temporal reasoning rather than global high-cost computation. Future work may integrate sparse long-range attention, dynamic actor-slot allocation, or memory-based identity mechanisms to extend TubeLite toward longer horizons and more crowded environments, while preserving its efficiency and modularity.

\vspace{-8pt}
\subsubsection{Acknowledgements}
This work was enabled by Archipelago Marine Research and NSERC Canada via the Alliance Grants program, and by support provided by the Digital Research Alliance of Canada.
\vspace{-8pt}

%
%
%
\bibliographystyle{splncs04}
\bibliography{references}

\end{document}